%% file: main.tex
\documentclass[10pt,twocolumn,letterpaper]{article}

\usepackage{iccv}      

\input{preamble}

\definecolor{iccvblue}{rgb}{0.21,0.49,0.74}
\usepackage[pagebackref,breaklinks,colorlinks,allcolors=iccvblue]{hyperref}

\def\paperID{3752} 
\def\confName{ICCV}
\def\confYear{2025}
\title{\mytitle}
\author{First Author\\
Institution1\\
Institution1 address\\
{\tt\small firstauthor@i1.org}
\and
Second Author\\
Institution2\\
First line of institution2 address\\
{\tt\small secondauthor@i2.org}
}

\author{
    Dejie Yang\textsuperscript{\rm 1},
    Zijing Zhao\textsuperscript{\rm 1},
    Yang Liu\textsuperscript{\rm 1,2}\thanks{Corresponding author}
    \\
    \textsuperscript{\rm 1} Wangxuan Institute of  Computer Technology, Peking University\\
    \textsuperscript{\rm 2} State Key Laboratory of General Artificial Intelligence, Peking University\\
    \{ydj,zijingzhao\}@stu.pku.edu.cn, yangliu@pku.edu.cn
}

\begin{document}
\maketitle
\input{sec/0_abstract}    
\input{sec/1_intro}
\input{sec/2_related}
\input{sec/3_method}
\input{sec/4_exp}
\input{sec/5_conclusion}

\noindent\textbf{Acknowledgements.}This work was supported by the grants from the National Natural Science Foundation of
China 62372014, Beijing Nova Program and Beijing Natural Science Foundation 4252040.

{
    \small
    \bibliographystyle{ieeenat_fullname}
    \bibliography{main}
}


\end{document}

%% file: preamble.tex
\usepackage{graphicx}	
\usepackage{amsmath}	
\usepackage{amssymb}	
\usepackage{booktabs}
\usepackage{times}
\usepackage{microtype}
\usepackage{epsfig}
\usepackage{caption}
\usepackage{float}
\usepackage{placeins}
\usepackage{color, colortbl}
\usepackage{stfloats}
\usepackage{enumitem}
\usepackage{tabularx}
\usepackage{xstring}
\usepackage{multirow}
\usepackage{xspace}
\usepackage{url}
\usepackage{subcaption}
\usepackage{graphbox}
\usepackage[hang,flushmargin]{footmisc}
\usepackage{pifont}       
\usepackage{bbding}       
\usepackage{fontawesome}  

\newcommand{\red}[1]{{\color{red}#1}}

\newcommand{\mytitle}[1]{{AR-VRM: Imitating Human Motions for Visual Robot Manipulation with Analogical Reasoning}#1}


%% file: sec/0_abstract.tex
\begin{abstract}
Visual Robot Manipulation (VRM) aims to enable a robot to follow natural language instructions based on robot states and visual observations, and therefore requires costly multi-modal data. To compensate for the deficiency of robot data, existing approaches have employed vision-language pretraining with large-scale data. However, they either utilize web data that differs from robotic tasks, or train the model in an implicit way (e.g., predicting future frames at the pixel level), thus showing limited generalization ability under insufficient robot data. In this paper, we propose to learn from large-scale human action video datasets  in an explicit way (i.e., imitating human actions from hand keypoints), introducing Visual Robot Manipulation with Analogical Reasoning (AR-VRM). To acquire action knowledge explicitly from human action videos, we propose a keypoint Vision-Language Model (VLM) pretraining scheme, enabling the VLM to learn human action knowledge  and directly predict human hand keypoints. During fine-tuning on robot data, to facilitate the robotic arm in imitating the action patterns of human motions, we first retrieve human action videos that perform similar manipulation tasks  and have similar historical observations , and then learn the Analogical Reasoning (AR) map between human hand keypoints and robot components. Taking advantage of focusing on action keypoints instead of irrelevant visual cues, our method achieves leading performance on the CALVIN benchmark {and real-world experiments}. In few-shot scenarios, our AR-VRM outperforms previous methods by large margins , underscoring the effectiveness of explicitly imitating human actions under data scarcity. Code available at \url{https://github.com/idejie/ar}.
\end{abstract}

%% file: sec/1_intro.tex
\section{Introduction}
\label{sec:intro}

Visual robot manipulation (VRM) is an essential task in the robotics field~\cite{kroemer2021review,fang2019survey,cong2021comprehensive}.
Based on current state and visual inputs, the robot is required to execute some actions according to human natural language instructions, including tasks such as object grasping, placement, and assembly~\cite{radosavovic2023real,brohan2023rt,brohan2022rt,wu2023unleashing}.
{Training such task requires multi-modal data, including paired images, natural language instructions, and robot action states under specific scenarios where robots work, which is costly and requires time-consuming human demonstrations {to} manipulate robots.
Therefore, the performance of VRM is often limited by the \textbf{scarcity of robot trajectories and annotations for training }\cite{brohan2022rt,wu2023unleashing}.}

\input{sec/figs/teaser}

To compensate for the deficiency of robot data, existing approaches utilize large-scale vision-language data pretraining and fine-tune the model under robot scenarios.  But, these pretraining data do not directly reflect object manipulation tasks, such as visual-question-and-answer datasets about animals and diets, and it is hard to provide relevant knowledge guidance for robot manipulation tasks.
Some studies \cite{nair2022r3m,wu2023unleashing} have utilized human action video datasets that are more similar to robotic manipulation tasks for pretraining, but they train the model in implicit ways, either by contrastive learning in feature space or by predicting future frames with pixel-level generative models.
Though acquiring human action knowledge, these approaches inevitably introduce irrelevant background information or pixel-level noise, which limits their performance on data-scarce VRM tasks.

In this paper, we propose to imitate human motions explicitly from a large-scale human action video dataset, training the model with human hand keypoints to learn from the motion itself, ignoring the irrelevant visual information.
To achieve this goal, two main challenges exist: 
(1) How to extract human action knowledge in terms of hand keypoints from large-scale human video datasets?
(2) How to train the robot, which differs from the human arm, to imitate human actions, i.e., building the correlation between the robotic components and human keypoints for the manipulation task?

To address these challenges in learning from human actions, we propose Visual Robot Manipulation with Analogical Reasoning (AR-VRM).
Human hand keypoints and robotic actions share underlying similarities, especially when it comes to manipulating objects.
Specifically, to extract action knowledge explicitly from a large-scale human video dataset, we propose a keypoint Visual-Language Model (VLM) pretraining scheme.
We use a large-scale egocentric human action video dataset Ego4D\cite{grauman2022ego4d} with common operation videos by human hands, which have similar manipulation tasks and environment views to robotic applications.
We detect human hand keypoints in instruction videos, and pretrain a VLM to predict the keypoints of future actions based on the current visual inputs and language instructions. 
{During the fine-tuning stage, to provide demonstrations under manipulation tasks, we first retrieve relevant human action videos that have similar history observations to the robot situation.
{Moreover}, to bridge the gap between robotic arm and human,  we  propose an Analogical Reasoning (AR) map between robotic arm components and human hand keypoints to learn geometric and functional correlation .}
Learning this correspondence helps to guide the robot to imitate human actions explicitly, e.g., the process of approaching the object and the grasping operation, to achieve manipulation tasks.

We perform extensive experiments on the challenging CALVIN benchmark.
Our AR-VRM achieves state-of-the-art performance under every experimental setting.
{Specifically, under 10\% robot training data as well as unseen scenarios, our method outperforms previous approaches by large margins, improving success rate from 40.0\% to 45.6\% under 10\% training data, and from 61.2\% to 65.9\% under unseen scenes.}
These results demonstrate that our idea of explicitly imitating human actions by keypoints has stronger generalization ability under limited robot data and under new scenarios. In addition, our model also outperforms previous SOTA methods on real robot experiments.

We conclude the contributions of our paper as follows:
(1) We propose AR-VRM, the first visual robot manipulation approach that explicitly imitates human actions in terms of hand keypoints from large-scale human action video datasets.
(2) We propose the keypoint vision language model pretraining scheme to extract action knowledge from human videos, and the analogical reasoning module to align robotic arms with human hand keypoints.
(3) Our method achieves state-of-the-art performance on CALVIN benchmark and real-world robot experiments, especially under limited robot training data, demonstrating the effectiveness and generalization ability.

%% file: sec/figs/teaser.tex
\begin{figure}
  \centering
  \begin{subfigure}{1.0\linewidth}
    \includegraphics[width=1\linewidth]{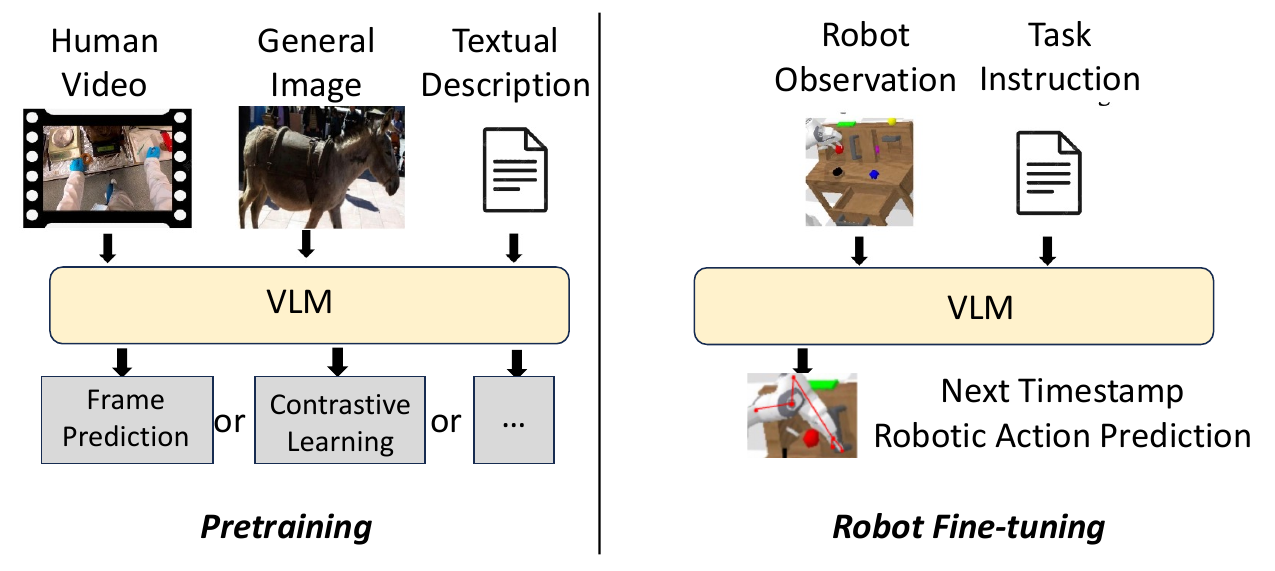}
    \caption{Previous methods: learning from large-scale data implicitly.}
    \label{fig:teaser-a}
  \end{subfigure}
 \hfill
  \begin{subfigure}{1.0\linewidth}
    \includegraphics[width=1\linewidth]{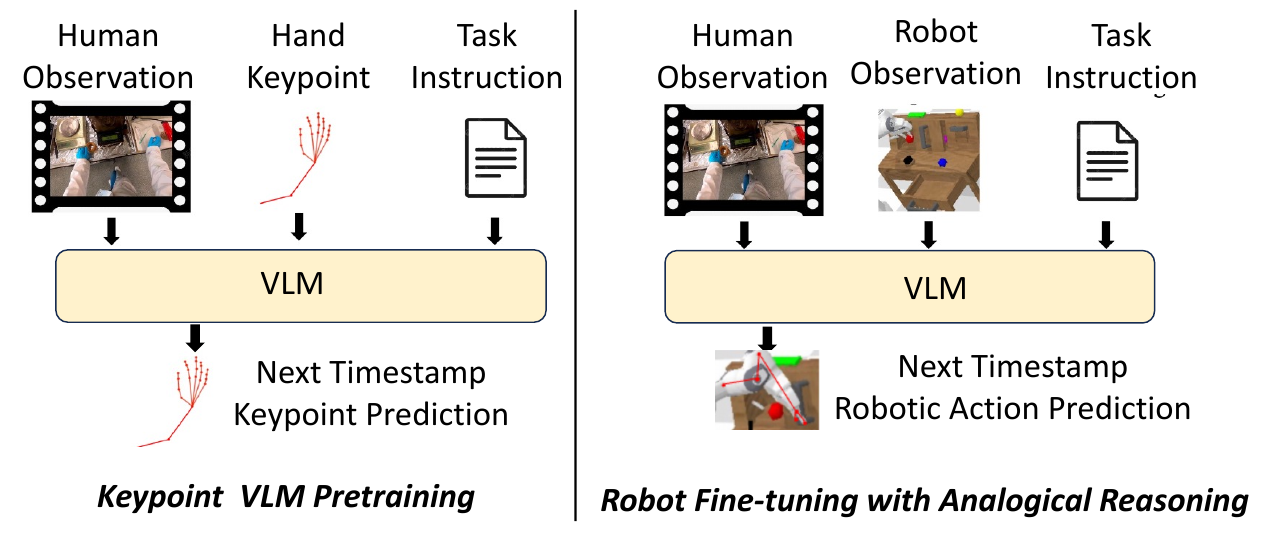}
    \caption{Our AR-VRM: explicitly learn from human action by hand keypoints.}
    \label{fig:teaser-b}
  \end{subfigure}  
   \label{fig:teaser}
  \caption{\textbf{Demonstration of the differences between our framework and previous methods:} we propose to learn from human actions explicitly by hand keypoints with analogical reasoning. }
\end{figure}

%% file: sec/2_related.tex
\section{Related Work}
\label{sec:related}

\subsection{Visual Robot Manipulation}

The visual robot manipulation task aims to train the robot to follow natural language instructions.
It is a flexible and intuitive way for non-expert humans to instruct the robot to execute tasks such as object grasping, placement, and assembly ~\cite{kroemer2021review,fang2019survey,cong2021comprehensive}.
Some pioneer approaches delved into diverse visual encoder pre-training methods, aiming to learn useful visual representations by masked image modeling ~\cite{radosavovic2023real,seo2023multi} and contrastive learning~\cite{nair2022r3m,jing2023exploring}, and reward signals for reinforcement learning~\cite{escontrela2024video}.
In recent years, some studies introduced large language models as the action planner \cite{song2023llmplanner,gao2024physically}.
ATM\cite{wen2023atm} proposes to imitate human actions for robot manipulation tasks, but requires paired human-robot action data under specific operation scenarios, limiting its generalization potential.
These methods all have limited performance due to the insufficient multimodal data, which is difficult to obtain and costly to collect.
In this paper, we employ large-scale human action video data for pretraining to compensate for the insufficiency of robot data.

\subsection{Vision Language Pretraining for Robotics}

Vision-Language Models (VLMs)  are pretrained on large-scale datasets with texts and images or videos, thus showing strong generalization ability on multiple downstream tasks, like video understanding\cite{TRM_2023_AAAI,tang2025video,yang2025planllm} ,  interaction grounding \cite{Lei_2024_CVPR,ting2024CMMP} and generation\cite{xusemantic,liu2025core4d},  and 3D scene understanding\cite{3DVQA_2024_AAAI,yang20243d}.
In the robotics field, some approaches introduce large-scale vision language pretraining to enhance robot intelligence.
\cite{brohan2023rt} employs Internet-scale web data with images and texts to pretrain the model for basic visual-language understanding, and fine-tune the model on robot data.
But the web data, for example, vision question answering data, differs from robot manipulation tasks, leading to limited performance gains.
\cite{nair2022r3m,wu2023unleashing} utilize human action video dataset, i.e., Ego4D\cite{grauman2022ego4d} which is closely related to manipulation task.
However, they train the robotic model in an implicit way, either by representation contrastive learning or by predicting pixel-level future frames, introducing irrelevant background information, and are unable to focus on the action itself.
In this paper, we propose to explicitly learn from human actions in terms of hand keypoints, enhancing the generalization ability under insufficient robot data.

%% file: sec/3_method.tex
\section{Method}
\label{sec:method}
\begin{figure*}
    \centering
    \includegraphics[width=\linewidth]{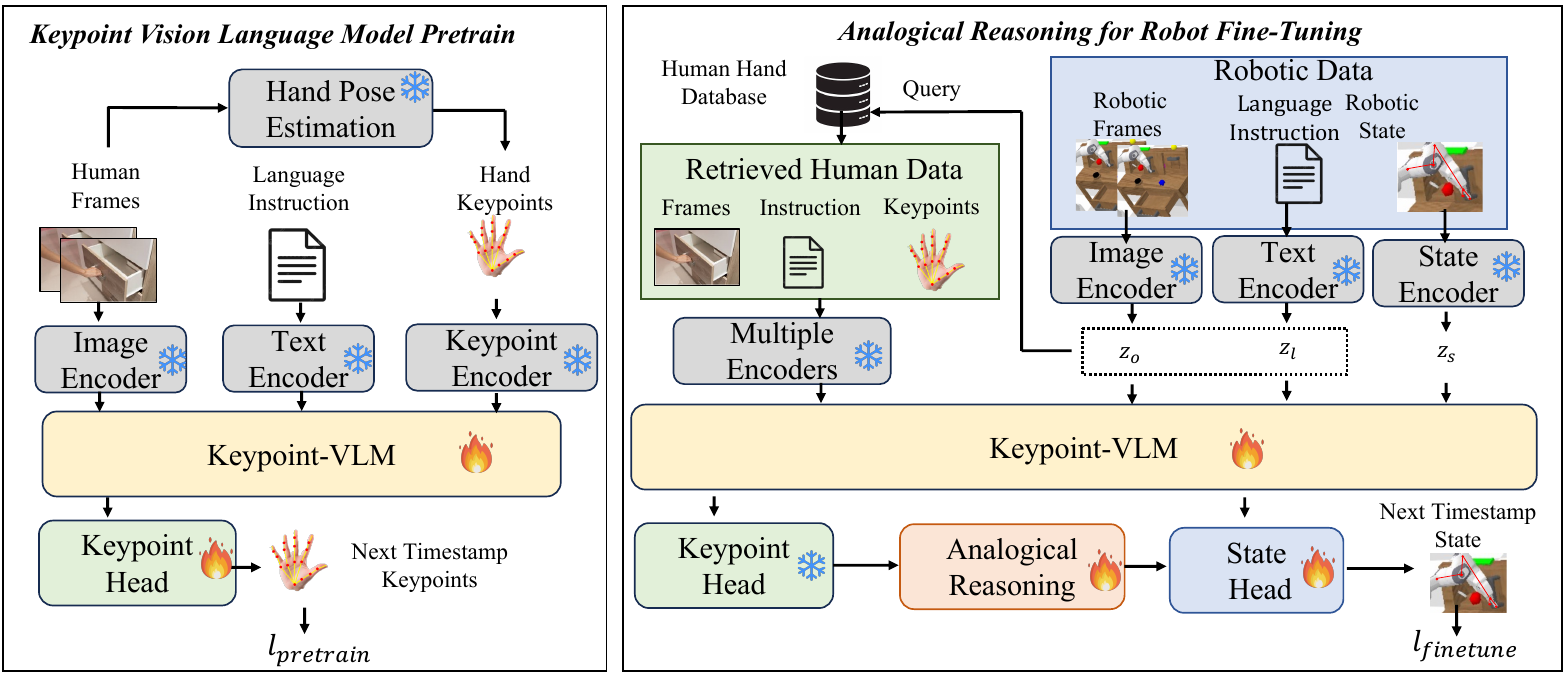}
    \caption{\textbf{AR-VRM: Visual Robot Manipulation  with Analogical Reasoning}.}
    
    \label{fig:frame}
\end{figure*}

\subsection{Problem Formulation and Method Overview}

We formulate the Visual Robot Manipulation (VRM) task as follows:
At timestep $t$, a robot model $R$ maps a language instruction $l$ and a sequence of history visual observations $o_{1:t}$ and robot states $s_{1:t}$ from the starting timestep to current timestep $t$ to a robot action $a_t$:
\begin{equation}
    R(l, o_{1:t}, s_{1:t}) \rightarrow a_t
\end{equation}
where $o_t \in \mathbb{R}^{H\times W\times 3}$ is the input image and $s_t=\{s_t^e, s_t^g\}$ is the corresponding robot state with the 6D pose of the robot end-effector $s_t^e \in \mathbb{R}^6$ and a binary state of the gripper $s_t^g \in \{0,1\}$ at timestep $t, t \in [1,T]$.
Action $a_t = \Delta(s_t) = s_{t+1} - s_t$ is the variation of state parameters.
The complete robot dataset $D_R = \{\tau^R_i\}_{i=1}^{|D_R|}$ contains paired language instructions, visual inputs, and robot states:
\begin{equation}
    \tau^R = \{l, o_1, s_1, o_2, s_2, ..., o_T, s_T\}
\end{equation}
These paired data with ground truth robot actions are difficult to obtain and costly to collect, thus have limited scale.

In this paper, we propose to employ vision language pretraining using large-scale human action videos and fine-tune the model on robot data, learning from explicit human action knowledge in terms of hand keypoints.
As overviewed in \Cref{fig:frame}, our framework  AR-VRM contains a keypoint Vision-Language Model (VLM) pretraining scheme, and an Analogical Reasoning (AR) module during fine-tuning. 

In the pretraining stage, we first extract human hand keypoints in the large-scale human action videos, formulating sequences of human action data, and then introduce a keypoint head for the VLM to directly predict future human body keypoints, acquiring human action knowledge and focusing on the explicit action keypoints.
In the fine-tuning stage, given the limited robot data, we first retrieve human actions that have a similar manipulation task as well as history observations from the human instructional video database, letting the pretrained keypoint VLM predict future actions, and introduce analogical reasoning to map the correspondence between the human hand keypoints and robot components.
In this way, the model learns from explicit human actions and can generalize well under insufficient robot data.

\subsection{Keypoint Vision-Language Model Pretraining}
We employ a large-scale human egocentric video dataset Ego4D\cite{grauman2022ego4d}.  
The video samples in Ego4D are human operations of massive-scale manipulation tasks, with similar environments and views to robotic applications.
In the human action video dataset $D_H = \{\tau^H_i\}_{i=1}^{|D_H|}$, a video of human hands accomplishing a task is provided:
\begin{equation}
    \tau^H = \{l, o_1, o_2, ..., o_T\}
\end{equation}
where $l$ is the language description of the task, and $o_t$ is the captured image at timestep $t$.
Unlike previous methods that learn implicitly from visual information, which inevitably contains irrelevant background information, we propose to explicitly learn from the human action itself with hand keypoints.
We employ an offline hand pose estimation model InterHand\cite{moon2020interhand2} to extract 3D hand keypoints in each video frame $k_t=\mathrm{InterHand}(o_t), k_t \in \mathbb{R}^K$, $K$ is the number of keypoints in human hands.
The preprocessed human action video dataset $D_{H^*} = \{\tau^{H^*}_i\}_{i=1}^{|D_{H^*}|}$ is:
\begin{equation}
    \tau^{H^*} = \{l, o_1, k_1, o_2, k_2 ..., o_T, k_T\}
\end{equation}
The keypoints extracted by \cite{moon2020interhand2} are under image coordinate system with  3D coordinates, which could be imitated by robotic arms, for example, mimicking the process of approaching an object and performing operations. 

Given the data from three different modalities, we pretrain a keypoint Vision-Language Model (VLM), enabling the VLM to understand and directly predict hand keypoints in a human action sequence.
For language instruction $l$, following \cite{shridhar2022cliport,shridhar2023perceiver,wu2023unleashing}, we use CLIP\cite{radford2021learning} text encoder to extract language embeddings, followed by a multi-layer perceptron (MLP) to project the embedding to dimension $d$:
\begin{equation}
    z_l = \mathrm{MLP}(\mathrm{CLIP_{text}}(l)), z_l \in \mathbb{R}^d
\end{equation}
For each visual input $o$, 
we employ a vision transformer (ViT) pretrained with MAE\cite{he2022masked} as the image encoder.
We take the output \texttt{CLS} token $z^{\texttt{CLS}}$ as the global representation and output patch tokens $z^{p_{1:V}}$ as local representations, resampled by a preceiver resampler (PR) \cite{jaegle2021perceiver} to reduce token numbers to $M$.
All the outputs are projected by MLP to dimension $d$: 
\begin{align}
    z^{\texttt{CLS}}, z^{p_{1:V}} &= \mathrm{ViT}(o_t) \\
    z_{o_t}^{\texttt{CLS}} &= \mathrm{MLP}(z^{\texttt{CLS}}), z_{o_t}^{\texttt{CLS}} \in \mathbb{R}^d \\
    z_{o_t}^{p_{1:M}} &= \mathrm{MLP}(\mathrm{PR}(z^{p_{1:V}})), z_{o_t}^{p} \in \mathbb{R}^d
\end{align}
For hand keypoints $k$, we employ HandFormer\cite{shamil2025utility} as the keypoint encoder, and also project the embedding to dimension $d$:
\begin{equation}
    z_{k_t} = \mathrm{MLP}(\mathrm{HandFormer}(k_t)), z_{k_t} \in \mathbb{R}^d
\end{equation}

With the aligned dimension of tokens from the three modalities $z_l, z_o = (z_{o}^{\texttt{CLS}}, z_{o}^{p_{1:M}}), z_k$, we concatenate them into a sequence of tokens and feed them to Transformer layers and perform next token prediction for pretraining.
Specifically, at timestep $t$, the model with self-attention layers $h^{\mathrm{Atten}}$ predict the keypoint token $\hat{z_k}$ based on the previous token sequences, and a keypoint prediction head (MLP) project the keypoint token back to keypoint vectors, and finally compute the mean squre error (MSE) with the ground truth keypoint. 
The pretraining loss can be formulated as:
\begin{align}\
    \hat{z_{k_t}} = h^{\mathrm{Atten}}(z_l, z_{o_1}, z_{k_1}, z_{o_2}, z_{k_2} ..., z_{o_{t-1}}, z_{k_{t-1}}) \\
    \mathcal{L}_{pretrain} = \sum_{t=1}^T \mathcal{L}_{MSE}\left(\mathrm{KeypointHead}(\hat{z_{k_t}}), k_t \right)
\end{align}

During training we fix the parameters of the CLIP text encoder and pretrained image encoder. 
With the keypoint VLM pretraining, we acquire human action knowledge from a large-scale human action video dataset and enable the VLM to explicitly predict action keypoints.

\subsection{Robot Fine-tuning with Analogical Reasoning}
With the keypoint VLM that understands action sequences by large-scale human data pretraining, we could then fine-tune the model on the robot data to accomplish the visual robot manipulation task.
In the robot dataset $D_R$, we extract the language and vision embeddings and produce tokens $z_l, z_o$ same as the pretraining stage.
For robot state $s = \{s^e, s^g\}$, we employ two MLPs $E^e, E^g$ to encode $s^e$ and $s^g$ respectively, and then project the encoded vectors into dimension $d$ by another MLP:
\begin{equation}
    z_s = \mathrm{MLP}(E^e(s^e), E^g(s^g)), z_s \in \mathbb{R}^d
\end{equation}
With the aligned dimension of tokens $z_l, z_o, z_s$, we feed them to the pretrained VLM to predict the state token $\hat{z_s}$.
Introducing the robot state head (MLP), we project the state token back to robot states and compute the MSE loss with the ground truth states:
\begin{align}
    \hat{z_{s_T}} = h^{\mathrm{Atten}}(z_l, z_{o_1}, z_{s_1}, z_{o_2}, z_{s_2} ..., z_{o_{T-1}}, z_{s_{T-1}}) \\
    \mathcal{L}_{state} = \mathcal{L}_{MSE}\left(\mathrm{StateHead}(\hat{z_{s_T}}), s_T \right)
\end{align}

In addition to utilizing the knowledge from pretrained weights of the VLM, we propose to explicitly learn from the human actions in terms of hand keypoints.
Specifically, given the robot data sample $\tau_R$, we first retrieve the human action videos from the large-scale database based on the language introduction and the visual frame features, with the similarity:
\begin{equation}
    sim(\tau^R, \tau^H) = cos(z_l^R, z_l^H) + \sum_{t}cos(z_{o_t}^R, z_{o_t}^H)
\end{equation}
where $cos(\cdot, \cdot)$ is the cosine similarity.
In this way, we could retrieve human action videos that have both similar manipulation tasks and similar visual observations with the robotic data sample.
We retrieve the top $J$ similar samples $\{\tau_j^H\}_{j=1}^J$.

After the forward pass of the human action samples $\tau_j^H$ and robot data sample $\tau^R$, the features of the last layer of keypoint head and robot state head $f_{k_j} \in \mathbb{R}^{K \times d}, f_s \in \mathbb{R}^{S \times d}$ serve as the representation of each keypoint node and robot state node, where $K$ is the number of keypoints of human hands and $S$ is the number of robot arms components.
We introduce a learnable analogical map matrix $m \in \mathbb{R}^{S \times K}$ to represent the mapping between the human hand keypoints and robot arm components, where each element represents the influence of the corresponding hand keypoint on the robot arm component.
We compute the imitated robot state features $f_{s_j}^*$ as follow:
\begin{equation}
    f_{s_j}^* = (1-\alpha)\cdot m \cdot f_{k_j} + \alpha \cdot f_s
\end{equation}
where $\alpha \in \mathbb{R}$ is also a learnable parameter to weight the overall influence of keypoint features.
We then employ a new linear layer for new robot states, and compute MSE as the analogical reasoning loss:
\begin{equation}
    \mathcal{L}_{AR} = \sum_{j=1}^J \mathcal{L}_{MSE}(\mathrm{Linear}(f_{s_{j,T}}^*), s_T)
\end{equation}

The overall fine-tuning loss is the weighted sum of the robot state loss and the analogical reasoning loss:
\begin{equation}
    \mathcal{L}_{finetune} = \mathcal{L}_{state} + \beta \cdot \mathcal{L}_{AR}
\end{equation}
where $\beta$ is the hyper-parameter.

\textbf{Discussion on the fine-tuning method design:}
Note that during fine-tuning, we fix the parameters of the keypoint encoder and keypoint head, and fine-tune the Transformer layers of the VLM.
Fixing the parameters of the keypoint encoder and keypoint head helps maintain the ability to encode and predict keypoint parameters that have already been accomplished in the pretraining stage.
By fine-tuning the Transformers of the VLM with human action video samples, we not only allow the keypoint features to guide the training of the robot state head, but also act as a data replay operation to prevent the VLM from over-fitting to the limited robotic data and forgetting the vision-language understanding and action prediction knowledge gained from pretraining.
Detailed experimental discussions will be in \Cref{sec:exp}.

%% file: sec/4_exp.tex
\section{Experiment}
\label{sec:exp}
\input{sec/tables/calvin}

\subsection{Dataset and Benchmark}

We conduct experiment on the challenging {\textit{long-horizon tasks benchmark} } CALVIN  \cite{mees2022calvin}.
CALVIN is a simulated language-conditioned robot manipulation benchmark that combines natural language conditioning, multi-modal high-dimensional inputs, 7-DOF continuous control, and long-horizon robotic object manipulation in both seen and unseen environments.
It contains 34 subtasks and 5 horizon evaluation sequences, providing a range of sensors commonly utilized for visuomotor control, and allows testing zero-shot generalization by leveraging a range of 4 manipulation environments and unseen language instructions.

Robot manipulation dataset requires costly paired multi-modal data, and thus has limited scale.
We employ a large-scale human action video dataset Ego4D \cite{grauman2022ego4d} for pretraining.
Ego4D contains massive-scale human-object interaction videos of 3,500 hours, each video containing a natural language annotation describing the human action.
Following \cite{wu2023unleashing}, we use a total of 800,000 video clips containing 8M frames for pretraining and as the human action video database for retrieval during fine-tuning. For real-world experiments, we use a plate with three objects: an orange, an apple, and a green jujube. We collected 1,200  moving-object demonstrations and   1,400 drawer-opening/closing trajectories.

\subsection{Implementation Details}

For network architecture, we use a pretrained CLIP\cite{radford2021learning} text encoder and a pretrained ViT-Base image encoder \cite{he2022masked} to extract language instruction tokens and visual input tokens.
For local image patch tokens, we use \cite{jaegle2021perceiver} to reduce the token numbers.
For human keypoints, we use the pretrained HandFormer\cite{shamil2025utility} as the encoder.
For robot states, we follow \cite{wu2023unleashing} to encode the robot arm and grasper parameters by MLPs.
We use a 12-layer transformer with causal attention mechanism and checkpoints loaded from \cite{radford2019language}.
During the pretraining stage, we sample uniformly spaced frames with 3 fps from video clips in the Ego4D dataset. And we set the learning rate to $1e-4$ with AdamW, training the model with a batch size of 512 on NVIDIA A800 GPUs with 100 epochs. During the fine-tuning phase, freezing multiple encoders that process human data and the keypoint prediction head serves to stabilize the model's existing performance while focusing updates on robotic manipulation prediction. And we adjust the learning rate to $1e-5$ and execute 50 epochs.

\subsection{Results Comparing with Other Methods}

\textbf{Experiment setup:}
We evaluate our method on the four different settings of the CALVIN benchmark, including full dataset multitask learning, unseen scene generalization, data-efficient few-shot learning, and unseen language generalization.
The training data includes ``A'', ``B'', ``C'' and ``D'' scenes with 34 specific instruction languages.
We use the success rate of sequentially completing 1, 2, 3, 4, and 5 tasks, and the average length of successfully completed tasks as the evaluation metrics following previous methods.

\textbf{Full dataset multitask learning:}
As shown in \Cref{tab:calvin} ``Experiment ABCD $\rightarrow$ D'' following \cite{wu2023unleashing}, we train the model on under 4 different scenes and evaluate it on a certain scene.
{Our method outperforms all the baseline methods, raising the average success rate improvement by +1.2\%, demonstrating the effectiveness of our design of introducing explicit human hand keypoints for robot manipulation.}

\textbf{Unseen scene generalization:}
As shown in \Cref{tab:calvin} ``Experiment ABC $\rightarrow$ D'', we train the model on under 3 scenes (``A'', ``B'', and ``C'') and evaluate it on scene ``D''.
Thanks to the replaying design of human videos during fine-tuning, our method further improves the performance by a large margin (Avg. Rate from 61.2\% to 65.9\%) compared with previous SOTA, indicating that our keypoint VLM pretraining and analogical reasoning have strong generalization ability across different scenes.

\input{sec/tables/real}
{\textbf{Real-robot experiments.} {(1)Object Transportation:}
\textit{Seen Objects.} The robot transported the three trained objects (orange, apple, jujube) in two disturbed scenes: with distractors (tomato, corn, yellow peach), and with an altered background (wooden board, bowl).
\textit{Unseen Instances.} Evaluated generalization to new instances of the same object categories (different orange, apple, jujube).
\textit{Unseen Categories.} Tested generalization to untrained categories (tomato, yellow peach).
Quantitative results (\Cref{tab:real}) show AR-VRM outperformed baselines (RT-1, MT-R3M, GR-1). Baselines often failed due to incorrect object selection/placement or collisions (e.g., with plate/desk). AR-VRM maintained high success rates for seen objects and showed minimal performance drop for unseen instances, highlighting strong in-category generalization.
(2){Articulated Manipulation:}
 It surpassed baselines significantly but exhibited two failure modes: incomplete drawer closure during closing tasks and missed handle engagement during opening tasks. Despite these, AR-VRM demonstrated superior robustness in articulated manipulation. 
}

\subsection{Ablation Studies and Qualitative Analysis}

\input{sec/tables/ablation}
\input{sec/tables/analogy}
\textbf{Effectiveness of our proposed modules:}
In \Cref{tab:ablation}, we show the results of ablating each of our proposed modules.
Line 1 represents training the VLM directly by robot data without any pretraining and guidance from human action videos.
Due to the lack of large-scale data, it has limited performance.
In line 2, with the help of human action video pretraining with hand keypoints, the VLM gains vision-language understanding knowledge, and the performance shows a significant gain.
Line 3 stands for the introduction of human action videos during fine-tuning by retrieval, but only training the separate keypoint head instead of the analogical mapping to guide the robot state prediction.
This method helps to maintain the knowledge learned by pretraining and prevents the over-fitting to small-scale robot data, thus showing performance gain compared with line 3, but still has room for improvement.
Line 4 stands for our complete design of keypoint VLM pretraining with large-scale data and fine-tuning with retrieved human action videos and analogical reasoning.
It achieves the best result, demonstrating the effectiveness of each of our proposed modules.

\textbf{Analysis of design choices of robot fine-tuning with analogical {reasoning:}}
{
As noted in \Cref{sec:method}, we freeze the keypoint encoder/head and fine-tune the VLM during analogical reasoning adaptation. Results in \Cref{tab:finetune} show:
Freezing the VLM (lines 1 vs 4, 2 vs 3) causes significant performance drops, confirming that VLM fine-tuning is critical for transferring human knowledge to robots.
Freezing keypoints (lines 1 vs 2, 3 vs 4) boosts performance, likely because stabilized keypoint features aid convergence of the analogical map, improving human-robot component alignment.}

\input{sec/tables/data}

\textbf{Date efficient few-shot learning:}
As shown in \Cref{tab:data}, we train the model with only 10\% of the already small dataset to train our model. Our method outperforms all the baseline methods, {raising the success rate of completing 5 tasks to 45.6\% and the average length of completed tasks to 2.28}. The results show that our analogical reasoning can provide a map from the diversity and complexity of actions in human keypoints to insufficient robot action, which can enable the robot to quickly learn how to cope with different tasks. 

\input{sec/tables/lang}
\textbf{Unseen instruction language generalization:}
As shown in \Cref{tab:lang}, we evaluate the model on the unseen instruction languages during training. 
Our method consistently achieves state-of-the-art, indicating that the relevant human action demonstration significantly help the robot generalize to multiple new instructional tasks.

\input{sec/figs/visualization}
\textbf{Visualization of our prediction results:}
In \Cref{fig:video}, we provide an example of robot manipulation task of ``grasp the blue block in drawer''.
Our method retrieves relevant human action videos such as ``pick up a tool/receipt/cloth/knife'' from the database.
By imitating the human action of picking up objects from the drawer, which contains process actions of approaching the drawer and grasping the object, the robot could successfully follow the language instruction to pick up the blue block in the drawer.

\input{sec/figs/analogy}

\textbf{Visualization of analogical map:}
In \Cref{fig:map}, we show an analogical map partial example from hand keypoints to a robotic arm, and the map is normalized in a column-wise manner and shows the correlation between the estimated hand keypoints and visible robotic arm components.
The coding of keypoint nodes and robotic arm component nodes is illustrated in (a) and (b), and in (c), the map elements that are highlighted by red circles indicate a strong relationship between the two nodes.
The grasper of the robot (code 0) is strongly related to the fingertips of human hands (code 4, 8, 12, 20) which operates the grasping in manipulation tasks, and the root segment of the robotic arm (code 3) relates to the human palm (code 0, 1, 5, 9, 17) deciding the direction of the movement.
This visualization indicates that our method learns a reasonable relationship by mapping the human action to guide the robotic arm.

%% file: sec/tables/calvin.tex
\begin{table*}[!ht]
    \centering
    \begin{tabular}{c|c|ccccc|cc}
        \toprule
        \multirow{2}{*}{\centering Method} & \multirow{2}{*}{\centering Experiment} & \multicolumn{5}{c|}{Success rate of tasks completed in a row} & \multirow{2}{*}{\centering Avg.Len.} & \multirow{2}{*}{\centering Avg. Rate}\\ 
         & & 1 & 2 & 3 & 4 & 5 &  \\
        \midrule
        MCIL\cite{lynch2021language} & \multirow{7}{*}{\centering ABCD$\rightarrow$D} & 0.373 & 0.027 & 0.002 & 0.000 & 0.000 & 0.40 &8.0\% \\ 
        RT-1\cite{brohan2022rt} & & 0.844 & 0.617 & 0.438 & 0.323 & 0.227 & 2.45&49.0\% \\ 
        HULC\cite{mees2022hulc} & & 0.889 & 0.733 & 0.587 & 0.475 & 0.383 & 3.07&61.3\% \\ 
        MT-R3M\cite{shamil2025utility} & & 0.752 & 0.527 & 0.375 & 0.258 & 0.163 & 2.07&41.5\% \\ 
        GR-1\cite{wu2023unleashing} & & {0.949} & {0.896} & {0.844} & 0.789 & 0.731 & {4.21}&84.2\% \\ 
        Ours & & \textbf{0.951} & \textbf{0.915} & \textbf{0.855} & \textbf{0.800} & \textbf{0.751} & \textbf{4.27}&\textbf{85.4\%} (\red{+1.2\%})\\ 
        \midrule
        MCIL\cite{lynch2021language} & \multirow{7}{*}{\centering ABC$\rightarrow$D} & 0.304 & 0.013 & 0.002 & 0.000 & 0.000 & 0.31&6.4\% \\ 
        RT-1\cite{lynch2021language} & & 0.533 & 0.222 & 0.094 & 0.038 & 0.013 & 0.90&18.0\% \\
        HULC\cite{mees2022hulc} & & 0.418 & 0.165 & 0.057 & 0.019 & 0.011 & 0.67&13.4\% \\
        MT-R3M\cite{shamil2025utility} & & 0.529 & 0.234 & 0.105 & 0.043 & 0.018 & 0.93 &18.6\%\\ 
        GR-1\cite{wu2023unleashing} & & 0.854 & 0.712 & 0.596 & 0.497 & 0.401 & 3.06 &61.2\% \\
        
        Ours & & \textbf{0.901} & \textbf{0.759} & \textbf{0.642} & \textbf{0.531} & \textbf{0.461} & \textbf{3.29}&\textbf{65.9\%} (\red{+4.7\%})\\
        \bottomrule
    \end{tabular}
    \caption{\textbf{Performance comparisons on CALVIN.} \textbf{Bold} represents the best results and \underline{underline} represents the second-best.}
    \label{tab:calvin}
\end{table*}

%% file: sec/tables/real.tex
\begin{table}[!h]
    \centering
    \begin{tabular}{c|ccc|c}
\toprule
         \multirow{3}{*}{\centering Approach} &   \multicolumn{3}{c|}{Transportation} & {\centering Articulated }\\
         & {Seen}  & \multicolumn{2}{c|}{Unseen}&   \multirow{2}{*}{Manipulation}\\
         &Obj.&Ins.&Cate.&\\\midrule
        RT-1\cite{brohan2022rt}  & 0.27 & 0.13 & 0.00 & {0.35}\\
        MT-R3M \cite{shamil2025utility} & 0.15 & 0.13 & 0.10 & {0.30}\\
        GR-1\cite{wu2023unleashing}  & 0.79 & 0.73 & 0.30 & {0.75}\\
        AR-VRM(Ours)& \textbf{0.95} & \textbf{0.91} & \textbf{0.53} & \textbf{0.82} \\
        \bottomrule
    \end{tabular}
    \caption{\textbf{Real robot experiment success rates}}
    \label{tab:real}
\end{table}

%% file: sec/tables/ablation.tex
\begin{table*}[!ht]
    \centering
    \begin{tabular}{ccc|ccccc|cc}
        \toprule
        \multirow{2}{*}{\centering Pretrain} & \multirow{2}{*}{\centering Retrieval} & \multirow{2}{*}{\centering AR}  & \multicolumn{5}{c|}{Success rate of tasks completed in a row} & \multirow{2}{*}{\centering Avg.Len.} &\multirow{2}{*}{\centering Avg. Rate}\\ 
         & & & 1 & 2 & 3 & 4 & 5 &  \\
        \midrule
        \ding{55} & \ding{55} & \ding{55} & 0.824 & 0.701 & 0.595 & 0.472 & 0.410 & 3.00 &60.0\%\\
        \ding{51} & \ding{55} & \ding{55} & 0.892 & 0.871 & 0.810 & 0.781 & 0.710 & 4.06 &81.3\%\\
        \ding{51} & \ding{51} & \ding{55} & 0.939 & 0.897 & 0.843 & 0.797 & 0.738 & 4.21&84.3\% \\
        \ding{51} & \ding{51} & \ding{51} & \textbf{0.951} & \textbf{0.915} & \textbf{0.855} & \textbf{0.800} & \textbf{0.751} & \textbf{4.27} &\textbf{85.4\%}\\ 
        \bottomrule
    \end{tabular}
    \caption{\textbf{Ablation studies on the effectiveness of our proposed modules.} ``Pretrain'', ``Retrieval'' and ``AR'' denotes keypoint VLM pretraining with large-scale human data, retrieving human action videos for fine-tuning and analogical reasoning, respectively.}
    \label{tab:ablation}
\end{table*}

%% file: sec/tables/analogy.tex
\begin{table*}[!ht]
    \centering
    \begin{tabular}{cc|ccccc|cc}
        \toprule
        \multirow{2}{*}{\centering Keypoint params} & \multirow{2}{*}{\centering VLM params} & \multicolumn{5}{c|}{Success rate of tasks completed in a row} & \multirow{2}{*}{\centering Avg.Len.}&\multirow{2}{*}{\centering Avg. Rate} \\ 
         & & 1 & 2 & 3 & 4 & 5 &  \\
        \midrule
        frozen & frozen & 0.942 & 0.845 & 0.803 & 0.741 & 0.598 & 3.93&78.6\% \\
        trainable & frozen & 0.931 & 0.832 & 0.788 & 0.726 & 0.576 & 3.85 & 77.1\%\\
        trainable & trainable & 0.939 & 0.847 & 0.821 & 0.772 & 0.730 & 4.11&82.2\% \\
        frozen & trainable & \textbf{0.951} & \textbf{0.915} & \textbf{0.855} & \textbf{0.800} & \textbf{0.751} & \textbf{4.27} &\textbf{85.4\%}\\  
        \bottomrule
    \end{tabular}
    \caption{\textbf{Ablation study on the design choices of robot fine-tuning with analogical reasoning.}}
    \label{tab:finetune}
\end{table*}

%% file: sec/tables/data.tex
\begin{table*}[!ht]
    \centering
    \begin{tabular}{c|c|ccccc|cc}
        \toprule
        \multirow{2}{*}{\centering Method}  &\multirow{2}{*}{\centering Data}  & \multicolumn{5}{c|}{Success rate of tasks completed in a row} & \multirow{2}{*}{\centering Avg.Len.} & \multirow{2}{*}{\centering Avg. Rate}\\ 
        & &  1 & 2 & 3 & 4 & 5 &  \\
        \midrule
        GR-1 &\multirow{2}{*}{\centering 100\%} & {0.949} & {0.896} & {0.844} & 0.789 & 0.731 & {4.21}&84.2\% \\ 
        Ours& &\textbf{0.951} & \textbf{0.915} & \textbf{0.855} & \textbf{0.800} & \textbf{0.751} & \textbf{4.27} &\textbf{85.4\%}\\ 
        \midrule
        GR-1 & \multirow{2}{*}{\centering 10\%}& 0.778 & 0.533 & 0.332 & 0.218 & 0.139 & 2.00&40.0\% \\
       
        Ours& & \textbf{0.809} & \textbf{0.589} & \textbf{0.381} & \textbf{0.287} & \textbf{0.213} & \textbf{2.28}&\textbf{45.6\%}\\
        \bottomrule
    \end{tabular}
    \caption{\textbf{Data efficient few-shot learning with only 10\% train data on ABCD$\rightarrow$D}.}
    \label{tab:data}
\end{table*}

%% file: sec/tables/lang.tex
\begin{table*}[!ht]
    \centering
    \begin{tabular}{c|c|ccccc|cc}
        \toprule
        \multirow{2}{*}{\centering Method}  &\multirow{2}{*}{\centering Lang}  & \multicolumn{5}{c|}{Success rate of tasks completed in a row} & \multirow{2}{*}{\centering Avg.Len.} & \multirow{2}{*}{\centering Avg. Rate}\\ 
         && 1 & 2 & 3 & 4 & 5 &  \\
        \midrule
        
        GR-1 &\multirow{2}{*}{\centering Seen} & {0.949} & {0.896} & {0.844} & 0.789 & 0.731 & {4.21}&84.2\% \\ 
        Ours& & \textbf{0.951} & \textbf{0.915} & \textbf{0.855} & \textbf{0.800} & \textbf{0.751} & \textbf{4.27} &\textbf{85.4\%}\\ 
        \midrule
        GR-1  &\multirow{2}{*}{\centering Unseen}  & 0.764 & 0.555 & 0.381 & {0.270} & {0.196} & {2.17}&43.3\% \\
        Ours & & \textbf{0.802} & \textbf{0.575} & \textbf{0.430} & \textbf{0.313} & \textbf{0.209} & \textbf{2.33}&\textbf{46.6\%}\\
        \bottomrule
    \end{tabular}
    \caption{
    \textbf{Unseen instruction language generalization on ABCD$\rightarrow$D}. }
    \label{tab:lang}
\end{table*}

%% file: sec/figs/visualization.tex
\begin{figure*}[!ht]
\vspace{-0.1cm}
    \centering

    \includegraphics[width=1\linewidth]{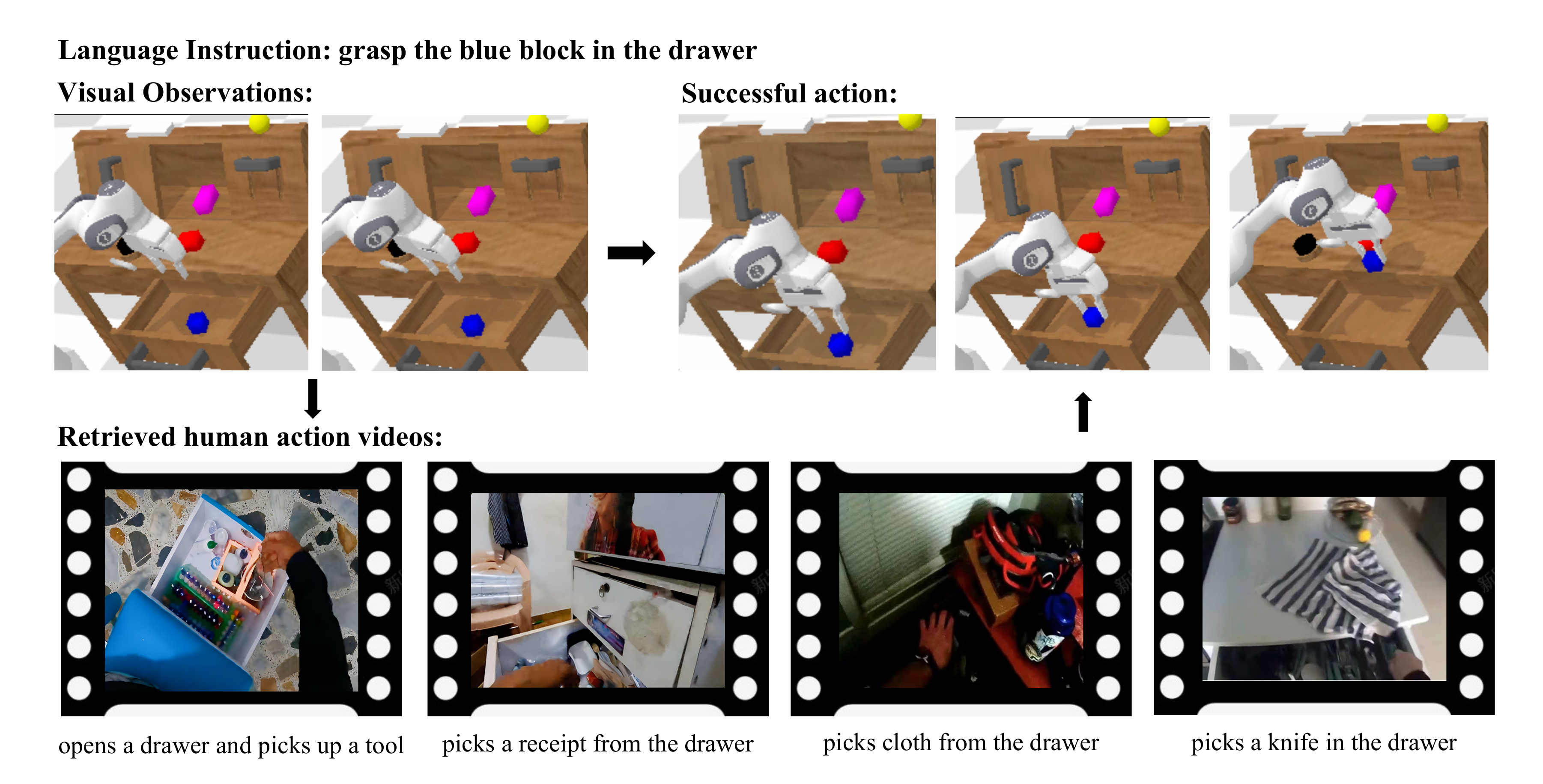}
    \caption{\textbf{Example of robot manipulation with our retrieved human action videos and action prediction result.}}
     \vspace{-0.2cm}
    \label{fig:video}
\end{figure*}

%% file: sec/figs/analogy.tex
\begin{figure*}[!ht]
    \centering
\includegraphics[width=0.8\linewidth]{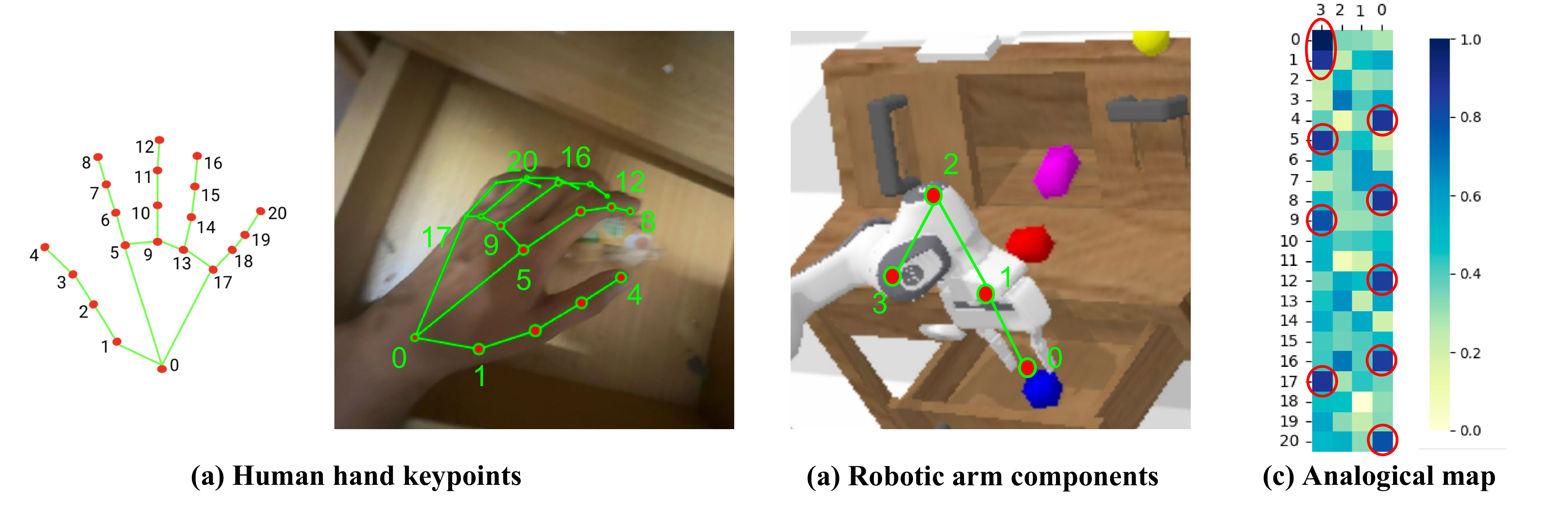}
    \caption{\textbf{Visualization of analogical mapping.} (a) Human hand keypoints code 0-20. (b) Robotic visible components code 0-3. (c) The learned analogical map. The highlighted map elements by red circles indicate a strong relationship between the two nodes.}
    \label{fig:map}
    \vspace{-0.5cm}
\end{figure*}

%% file: sec/5_conclusion.tex
\section{Conclusion}
\label{sec:con}
To explore key information in human interaction processes from videos to guide robot manipulation,  we introduce a Visual Robot Manipulation with Analogical Reasoning (AR-VRM), leveraging large-scale human action video datasets to explicitly learn action knowledge through hand keypoints. We use a keypoint pretraining scheme to enhance generalization and performance. With Analogical Reasoning to map human hand keypoints to robot components, our method achieves SOTA results on the CALVIN and real robot experiments, especially in few-shot data scenarios, demonstrating our effectiveness and generalization. 